\documentclass{article}

\usepackage[preprint]{neurips_2025}

\usepackage[utf8]{inputenc} 
\usepackage[T1]{fontenc}    
\usepackage{hyperref}       
\usepackage{url}            
\usepackage{booktabs}       
\usepackage{amsfonts}       
\usepackage{nicefrac}       
\usepackage{microtype}      
\usepackage{xcolor}         

\usepackage{graphicx}
\usepackage{amsmath}
\usepackage{tcolorbox}
\usepackage{booktabs}
\usepackage{tabularx}
\usepackage{multirow}
\usepackage{xspace}

\newcommand{\parboxc}[1]{\parbox[c]{\hsize}{\vspace{2mm}#1\vspace{2mm}}}
\usepackage{longtable}

\newcommand{\generator}{\textbf{G}$_{\theta_1}$\xspace}

\newcommand{\student}{\textbf{S}$_{\theta_2}$\xspace}
\newcommand{\wset}{\mathcal{W}}
\newcommand{\archive}{\mathcal{A}}

\newcommand{\ssFunction}{\textbf{C}}
\newcommand{\upFunction}{\textbf{U}}

\newcommand{\solverate}{\textsc{SolveRate}}

\newcommand{\methodlong}[0]{Synthetic Problem Generation for Reasoning via Quality-Diversity Algorithms}
\newcommand{\method}[0]{\textsc{SPARQ}}

\newcommand{\ppara}[1]{\vspace{-0.3cm}\paragraph{#1}}

\title{SPARQ: Synthetic Problem Generation for Reasoning via Quality-Diversity Algorithms}

\author{%
  Alex Havrilla$^{1,3,*}$  \\
  \And
   Edward Hughes$^{2}$ \\
   \And
   Mikayel Samvelyan$^{2}$ \\
   \And
   Jacob Abernethy$^{1,3}$ \\
}

\begin{document}

\maketitle

\renewcommand{\thefootnote}{}
\footnotetext{$^*$Work done during an internship at Google Research. Correspondence to \texttt{ahavrilla3@gatech.edu}.}

\vspace{-0.75cm}
\begin{center}
   Google Research${^1}$\quad Google DeepMind${^2}$\quad Georgia Tech${^3}$
\end{center}
\vspace{1cm}

\begin{abstract}
  Large language model (LLM) driven synthetic data generation has emerged as a powerful method for improving model reasoning capabilities. However, most methods either distill large state-of-the-art models into small students or use natural ground-truth problem statements to guarantee problem statement quality. This limits the scalability of these approaches to more complex and diverse problem domains. To address this, we present \methodlong{} (\method{}), a novel approach for generating high-quality and diverse synthetic math problem and solution pairs using only a single model (\texttt{Gemma-2-9b}) by measuring a problem's \textit{solve-rate}: a proxy for problem difficulty. Starting from a seed dataset of 7.5K samples, we generate over 20 million new problem-solution pairs.
  We show that filtering the generated data by difficulty and then fine-tuning the same model on the resulting data improves relative model performance by up to 24\%. Additionally, we conduct ablations studying the impact of synthetic data quantity, quality and diversity on model generalization. We find that higher quality, as measured by problem difficulty, facilitates better in-distribution performance. Further, while generating diverse synthetic data does not as strongly benefit in-distribution performance, filtering for more diverse data facilitates more robust OOD generalization. We also confirm the existence of model and data scaling laws for synthetically generated problems, which positively benefit downstream model generalization.
\end{abstract}

\section{Introduction}
\label{sec:intro}

The quantity of high-quality problem statements is one of the most impactful factors affecting the downstream performance of both supervised fine-tuned (SFT) and RL fine-tuned models \citep{toshniwal2024openmathinstruct118millionmath,singh2024humandatascalingselftraining}. However, the vast majority of synthetic data generation approaches either (1) generate only new solutions to a fixed problem set \citep{toshniwal2024openmathinstruct118millionmath, toshniwal2024openmathinstruct2acceleratingaimath,havrilla2024teachinglargelanguagemodels} or (2) generate new problems using a large state-of-the-art oracle without carefully filtering the resulting problem data for quality and correctness \citep{yue2023mammothbuildingmathgeneralist, yu2024metamathbootstrapmathematicalquestions}. These approaches sidestep a key difficulty in problem generation: the lack of a ground truth signal which can be used to filter out illogical/invalid problems. As a result, these approaches are not scalable to generating new problem-solution pairs whose difficulty exceeds the capabilities of existing SOTA models. Additionally, the restriction to natural data limits problem-solution diversity, especially in the increasingly complex domains in which models are applied. 

In this work we present \methodlong{} (\method{}), a novel synthetic data generation algorithm producing high-quality and diverse problem-solution pairs using a single student model \student (\texttt{Gemma-2-9b} \citep{gemmateam2024gemma2improvingopen}) and seed dataset $\mathcal{D}$. The key idea behind \method{}  is to use monte-carlo rollouts from \student to estimate the \textit{solve-rate} (i.e., "difficulty") of a problem-solution pair  $(Q, A)$ for \student. We then use the solve-rate as a proxy for problem quality, allowing us to filter out low-quality generations which are either too hard/impossible or too easy. 
We demonstrate that generating and filtering data based on this quality score significantly improves the pass@1 accuracy of the MATH~\citep{hendrycks2021measuringmathematicalproblemsolving} SFT \texttt{Gemma-2-9b} student from 38\% to 47\%. An ablation confirms that increased average train problem quality correlates well with improved downstream performance.

We then investigate the role of problem diversity in problem generation and downstream generalization using \method{}. Problem diversity is measured first by annotating problem-solution pairs $(Q, A)$ with the skill/technique set used in solving $Q$. For example, $Q$ may require a combination of the $\texttt{pigeonhole principle}$ and $\texttt{algebra}$ in the correct solution. Representing problems with their most relevant skills allows us to measure the overall diversity of a set of problems by examining the \textit{coverage} (number of unique skills) and \textit{redundancy} (i.e., number of repeated skills). Given a fixed sample budget $N$, we find increasing problem diversity does not improve \textbf{in-distribution} generalization relative to a randomly selected training data baseline. However, we find models trained on more diverse synthetic data generalize better to out-of-distribution (\textbf{OOD}) tasks as the amount of compute used to solve the task increases. 

In summary, we make the following contributions:

\begin{enumerate}
    \item We introduce the \textit{solve-rate} of a problem $Q$ with respect to \student as an effective way of measuring and filtering by synthetic problem quality.
    \item We present \method{}, a new approach for synthetic problem generation that directly optimizes for the data quality and diversity. Training on the resulting data leads to an absolute downstream improvement over the baseline of 9\%.
    \item We carefully ablate the impact of quality and diversity of synthetic data generated with \method{} on model generalization, revealing increased quality to correlate well with better in-distribution generalization and diversity to benefit OOD generalization.
\end{enumerate}

\section{\methodlong{}}
\label{sec:methods}

\begin{figure}
    \includegraphics[width=\linewidth]{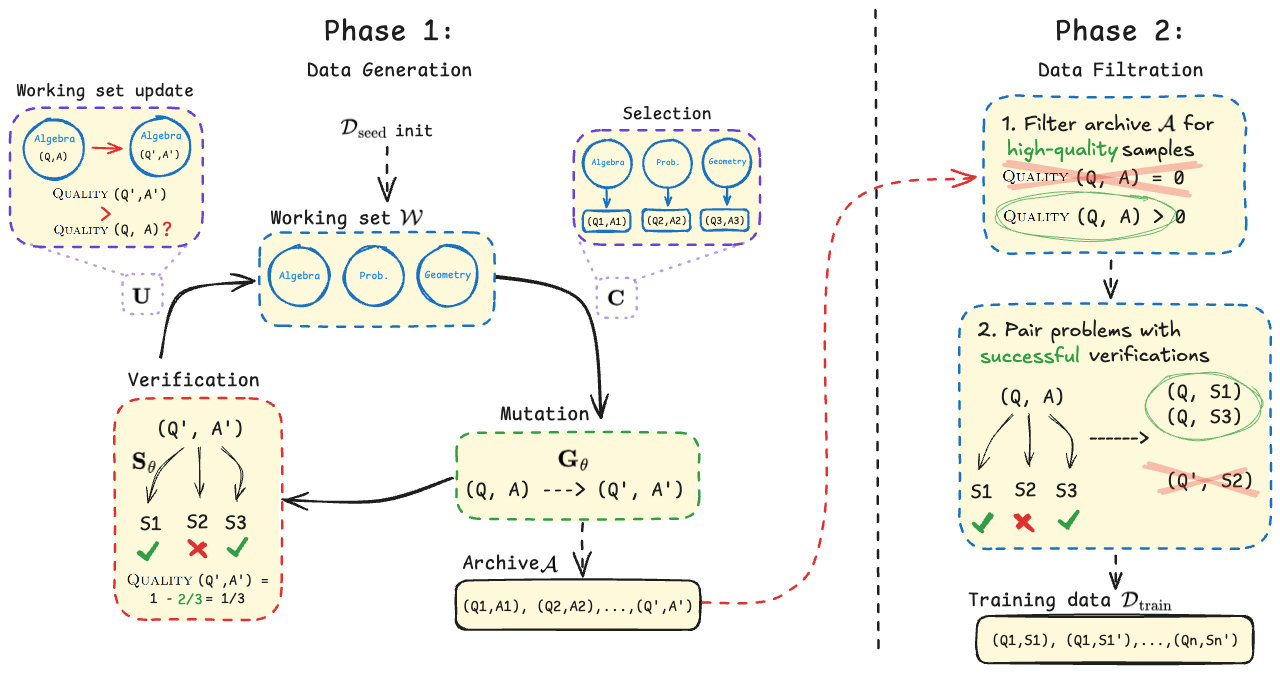}
    \caption{Diagram of the synthetic training data generation pipeline with \method{}. \textbf{Phase 1:} Data generation begins with $\mathcal{D}_{\text{seed}}$ initializing the working set. Samples $(Q, A)$ are iteratively selected from the working set $\mathcal{W}$ via the selection distribution \textbf{C} and mutated via the generator \generator. All new mutations are stored in the archive $\mathcal{A}$. The quality of new samples $(Q', A')$ is assessed via the difficulty student model \student has solving for $A'$ given $Q'$. The working set update function $\textbf{U}$ then updates $\mathcal{W}$ with new sample $(Q', A')$. $\textbf{Phase 2:}$ After data generation, data filtration proceeds by removing all low-quality samples from the archive $\mathcal{A}$. Each remaining question $Q$ is then paired with its successful verifications $S_k$ forming synthetic training tuples $(Q, S_k)$.}
    \label{fig:qd_problem_diagram}
\end{figure}

\ppara{Setup and notation} Let $\theta_1 \in \mathbb{R}^{d_1}, \theta_2 \in \mathbb{R}^{d_2}$ be optionally distinct model parameters. Suppose we are given a generating language model \generator and student model \student. Fix a tokenized vocabulary $\mathcal{V} = \{1, ..., V\} \subseteq \mathbb{N}$ for some $V \in \mathbb{N}$. Let the task $\tau$ be a distribution of problem-solution pairs $(Q, A) \in \mathcal{V}^{L_Q} \times \mathcal{V}^{L_A}$ where $L_Q, L_A \in \mathbb{N}$ are the maximum context lengths of problems and solutions respectively. Let $\mathcal{D}$ be a seed dataset of $n \in \mathbb{N}$ problem-solution pairs $(Q_i, A_i) \sim \tau$.

\ppara{A high-level synthetic data generation recipe} Given $\mathcal{D}$ we initialize a \textit{working set} set $\wset^{(0)}$ of mutation candidates meant to hold high-quality samples for future mutation. We also initialize an \textit{archive} $\archive^{(0)}$ which accumulates all generated data regardless of quality. Our synthetic data generation algorithm then proceeds in two phases: data generation and data filtration. The \textbf{data generation} phase proceeds iteratively at round $t$ by sampling a batch of problem-solution pairs $(Q_i^{(t)}, A_i^{(t)}) \sim \textbf{C}$ of size $b \in \mathbb{N}$ from the working set $\wset^{(t)}$ using the mutation selection distribution $\textbf{C}^{(t)} \in \Delta(\wset^{(t)})$. Mutations $({Q'}_i^{(t)}, {A'}_i^{(t)}) \sim $ \generator $(\cdot | (Q_i, A_i))$ are then sampled using the generator \generator. We call $(Q_i^{(t)}, A_i^{(t)})$ the parent and $({Q'}_i^{(t)}, {A'}_i^{(t)})$ the child or mutation of $(Q_i^{(t)}, A_i^{(t)})$. We then measure the $\textsc{Quality}$ of the new pairs $({Q'}_i^{(t)}, {A'}_i^{(t)})$ and update the working set $\wset^{(t)} \to \wset^{(t+1)}$ via the update function $\textbf{U}^{(t)}$. The update function $\textbf{U}^{(t)}\left( \wset^{(t)}, \{({Q'}_1^{(t)}, {A'}_1^{(t)}),...,({Q'}_b^{(t)}, {A'}_b^{(t)})\} \right) \to \wset^{(t+1)}$ inserts the high-quality subset of new mutations into $\wset^{(t)}$ and removes excess low-quality data in $\wset^{(t)}$ to produce $\wset^{(t+1)}$. Finally, we take the union of new mutations $\{({Q'}_1^{(t)}, {A'}_1^{(t)}),...,({Q'}_b^{(t)}, {A'}_b^{(t)})\} \bigcup \archive^{(t)}$ to produce $\archive^{(t+1)}$. Note: when it is clear we will suppress the notational dependence on the round $t$.

Once all rounds of data generation conclude, the \textbf{data filtration} phase begins. During this phase, every synthetic pair $(Q, A) \in \archive$ in the archive $\archive$ resulting from the data generation phase is filtered by evaluating whether $\textsc{Quality}((Q, A)) > 0$. The remaining data is used to construct a new training dataset $\mathcal{D}'$. Figure \ref{fig:qd_problem_diagram} shows a diagram of the entire pipeline.

\ppara{Measuring data quality} Given an arbitrary problem-solution pair $(Q, A)$ the key idea behind our approach is to use the average score the student model \student receives when solving $Q$. We call this quantity the \textit{solve-rate} of \student on $Q$ which is computed via $K$ many rollouts $S_1, \ldots, S_K \sim$ \student $(\cdot | Q)$ i.e.

\vspace{-1cm}
\begin{align*}
    \solverate((Q,A), \textbf{S}_\theta) := \frac 1 K \sum_{k=1}^K \textup{is\_correct}(Q, A, S_k)
\end{align*} where $\textup{is\_correct}(Q, A, S_k)$ is $1$ if $S_i$ agrees with the intended solution $A$ (in practice we check for the same numerical final solution) else $0$; it is for this reason we call each rollout $S_k \sim $ \student $(\cdot |Q)$ a \textit{verification} of $Q$. Intuitively, the solve-rate of $(Q, A)$ measures the difficulty of $Q$ for student \student. We use the solve-rate of a problem $Q$ to define a quality score for $Q$ as

\vspace{-0.5cm}
\begin{align*}
    \textsc{Quality}((Q,A), \textbf{S}_\theta) = \begin{cases}
        1 - \solverate((Q, A), \textbf{S}_\theta), & T_l \leq \solverate((Q, A), \textbf{S}_\theta) \leq T_u \\
        0 & \textup{otherwise}
    \end{cases}
\end{align*} which defines the quality of a problem as one minus its solve-rate for problems with a solve-rate between $T_l$ and $T_u$ and $0$ otherwise. We introduce the thresholds $0 < T_l < T_u < 1$ for three reasons: (a) to remove impossible problems i.e. those which are never verified by \student; (b) to attempt to remove problems which are sometimes verified by \student but contain reasoning errors in their intended solution $A$; c) to remove trivial pairs too easy for the student. Note: we call such pairs $(Q, A)$ containing a reasoning error in $A$ invalid. 

Evaluating the quality of a pair $(Q, A)$ with its solve-rate comes with the added benefit that successful verifications $S_k$ of $Q$ produced by \student can be added to the training data for free. Thus our final training samples are of the form $(Q, S)$ where $Q$ is a high-quality problem generated by \generator and $S$ is a successful verification of $Q$ via \student. Since we will be training \student with this new $\mathcal{D}_{\textup{train}}'$ \textbf{this makes all training solutions entirely self-generated}.

\ppara{Measuring data diversity} In addition to data quality, we are also interested in the \textit{diversity} of synthetically generated data. Abstractly, the diversity measure $\textsc{Div} : \mathcal{P}(\textup{support}(\tau)) \to \mathbb{R}$ is a function from sample subsets of the task $\tau$ to $\mathbb{R}$ measuring some notion of the subset's \textit{coverage} of $\tau$ or \textit{self-redundancy}. Concretely, we propose to measure the diversity of a set of problem-solution pairs $\{(Q_1, A_1),...,(Q_n, A_n)\}$ via their representation as \textit{skill-sets}. Informally, the \textit{skill-set} of a pair $(Q, A)$ is the alphabetically ordered k-tuple of semantic skills $\Phi((Q, A)) = \phi = (\phi_1, ..., \phi_k)$ which are used in $A$ to solve $Q$. For example, $Q$ might require a combination of  \textit{algebra} ($\phi_1$) and \textit{polynomial factoring} ($\phi_2$) for a correct solution. This representation enables a number of diversity measures. For this work we simply count the number of unique skills in a problem set $\{(Q_1, A_1),..., (Q_n, A_n)\}$, i.e.

\vspace{-0.5cm}
\begin{align*}
    \textsc{Div}(\{(Q_1, A_1),..., (Q_n, A_n)\}) = \large| \{\phi_j : \phi = \Phi( (Q_i, A_i)), 1 \leq j \leq k\} \large|
\end{align*} We limit the set of all possible skills as $\{\phi_1,...,\phi_M\}$ with fixed size $M \in \mathbb{M}$.





\subsection{Variations on the high-level data generation recipe}
\label{subsec:algo_variations}

In Section \ref{sec:methods} we defined a novel high-level data generation pipeline for producing high-quality synthetic data. However, the choices of sample selection distribution $\ssFunction$ and working set update function $\upFunction$ are left undefined. Here we detail four different implementations of the above pipeline with different choices for $\ssFunction$ and $\upFunction$ affecting the resulting quality and diversity of generated data.

\ppara{Static uniform data generation} Our simplest method is \textit{static uniform data generation}. Mutation parent samples are selected uniformly at random from the working set $\wset$,  that is, the sample selection function $\ssFunction$ is uniform. The update function $\upFunction$ keeps the working set $\wset$ fixed. As a result, only samples from the seed-dataset are sampled for mutation.  

\ppara{Static diverse data generation} We modify the above implementation slightly by partitioning the working set $\wset$ into equivalence classes $[Q, A]_\phi$ by identifying samples with the same skill-sets $\phi$. Formally, $[Q, A]_\phi = \{(Q, A) : \Phi((Q, A)) = \phi\}$. In a slight abuse of notation, we also call each equivalence class a \textit{skill-set}. $\ssFunction$ then samples uniformly first over skill-sets and then over class elements with the skill-set.

\ppara{Dynamic uniform data generation} We again slightly modify the static uniform generation procedure by iteratively updating the working set $\wset$ to contain the $T \in \mathbb{N}$ highest-quality samples in the archive $\archive$. This is done by inserting high-quality mutations and removing the lowest-quality samples from $\wset^{(t)}$ to produce $\wset^{(t+1)}$. Samples for mutation are selected uniformly from $\wset$ as in static uniform generation.

\ppara{Dynamic diverse data generation} Finally, inspired by algorithms from the quality-diversity literature \citep{mouret2015illuminatingsearchspacesmapping, pourcel2024acesgeneratingdiverseprogramming, samvelyan2024rainbowteamingopenendedgeneration}, we combine the modifications proposed in the above two methods to propose \textit{dynamic diverse data generation}. As with static diverse generation, we partition the working set $\wset^{(t)}$ into skill-set based equivalence classes which are sampled uniformly to produce mutation parents. After mutation, we assign the skill-sets $\phi^1,...,\phi^n$ to new sample mutations and insert high-quality samples $(Q, A)$ into their corresponding skill-set class $[Q, A]_{\Phi((Q, A))}$. For each class we enforce a uniform population limit $T_\phi = T$ by removing the lowest-quality samples in a class when new high-quality samples are added. Intuitively, the goal of this quality-diversity driven data generation algorithm is to generate maximally diverse, high-quality problems with as many unique skill-sets and difficulty levels as possible.

\section{Experiments}
\label{sec:experiments}

\ppara{Setup} We apply \method{} to improve the math reasoning abilities of LLMs. As our seed dataset $\mathcal{D}$ we use the 7.5K train set from MATH \citep{hendrycks2021measuringmathematicalproblemsolving}. We utilize the Gemma-2 model series \citep{gemmateam2024gemma2improvingopen}, always taking \texttt{Gemma-2-9b} as our student model \student. To produce a strong student model \student we fine-tune the pre-trained \texttt{Gemma-2-9b} on $\mathcal{D}$ for $3$ epochs. Unless otherwise specified, we use the instruction tuned \texttt{Gemma-2-27b-it} as our problem generator \generator (see Subsection \ref{subsec:self-improvement} for recursive self-improvement results). $K = 16$ rollouts are used to compute the solve-rate. A solve-rate thresholding between $T_l = 0.1$ and $T_u = 0.9$ is used. In the diversity-driven methods, we use the $M = 100$ most commonly occurring skills in $\mathcal{D}$ with combinations of size at most $k = 3$ to measure diversity. \texttt{Gemma-2-9b-it} is used to identify the problem skill-sets. Prompts for problem generation and skill classification are shared in Appendix \ref{sec:prompts}.

We run each data generation method with a mutation batch size of 64 for a maximum of 5K steps. This results in an archive $\archive$ with 320K synthetic problem-solution pairs. Note: because each problem requires $K = 16$ verifications for the quality evaluation, a total of 5 million synthetic solutions are generated (per method) in addition to problem statements and skill classifications. We find that the vast majority of generated problems are too difficult for \student to solve, resulting in a $\textsc{SolveRate}$ of 0. As a result, the average final size of a synthetically generated training dataset $\mathcal{D}_{\text{train}}$ comes out to around 80K unique problems with $\textsc{Quality}(Q) > 0$ and a corresponding 500K $(Q, S)$ problem-verification pairs. Figure \ref{fig:score_dists} in the Appendix shows a histogram of the distribution of problem solve-rates. See Appendix \ref{sec:hparams} for a description of training hyperparameters.

\subsection{Main Results for Problem Generation}
\label{subsec:main_results}

\begin{figure}
    \centering
    \includegraphics[width=0.48\linewidth]{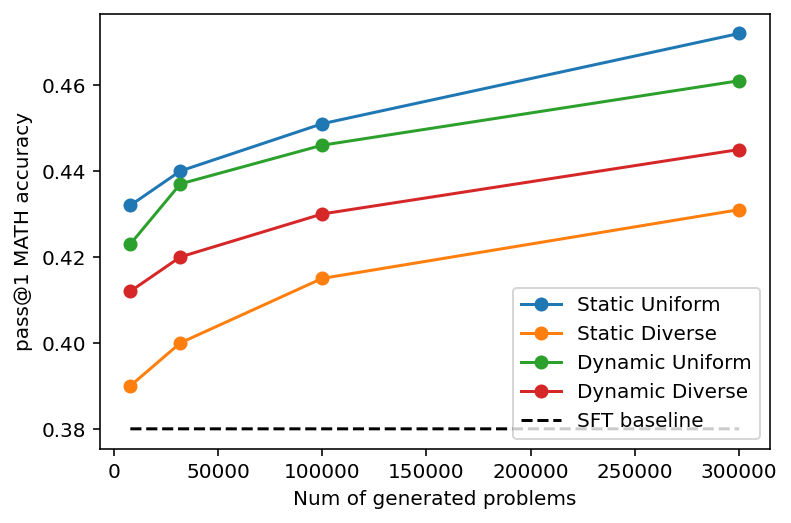}
    \includegraphics[width=0.48\linewidth]{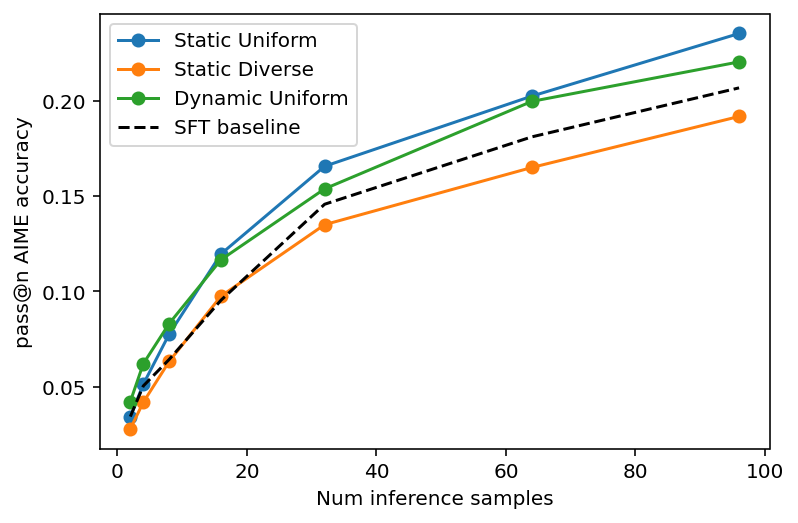}
    \caption{Performance of the downstream \texttt{Gemma-2-9B} models trained on synthetic data generated with \method{}. \textbf{Left:} \textbf{In-distribution} performance of  students on MATH test. On the x-axis is the number of synthetically generated problem-solution pairs. On the y-axis is MATH performance. Each curve plots the performance of a different data generation strategy. \textbf{Right:} \textbf{OOD} performance of downstream students on the AIME benchmark after training with 100K generated problems. On the x-axis is the number of inference-time solutions. On the y-axis is AIME pass@n accuracy.}
    \label{fig:scaling_results}
\end{figure}

\ppara{Self-synthetically generated  data significantly improves over the SFT baseline} Figure \ref{fig:scaling_results} illustrates the in-distrbution and OOD performance of downstream models trained on the resulting data from each variant of \method{}. We find training on data generated by every method improves over the SFT baseline. In particular, the static uniform method improves by up to an absolute 9\%, increasing from 38\% to 47\%. Downstream performance also benefits from scaling the size of the problem set: tripling the number of generated problems leads to a roughly 1.5\% performance increase. When scaling the inference compute during OOD evaluation, the static uniform method improves over the SFT baseline from 20\% to 25\% (with $K = 96$ inference samples). The OOD performance improvement can be seen at all inference compute budgets, increasing as the budget increases. This demonstrates the problems generated by \generator do not overfit to the initial seed dataset $\mathcal{D}$, allowing for an OOD improvement.

\begin{figure}
    \centering
    \includegraphics[width=0.7\linewidth]{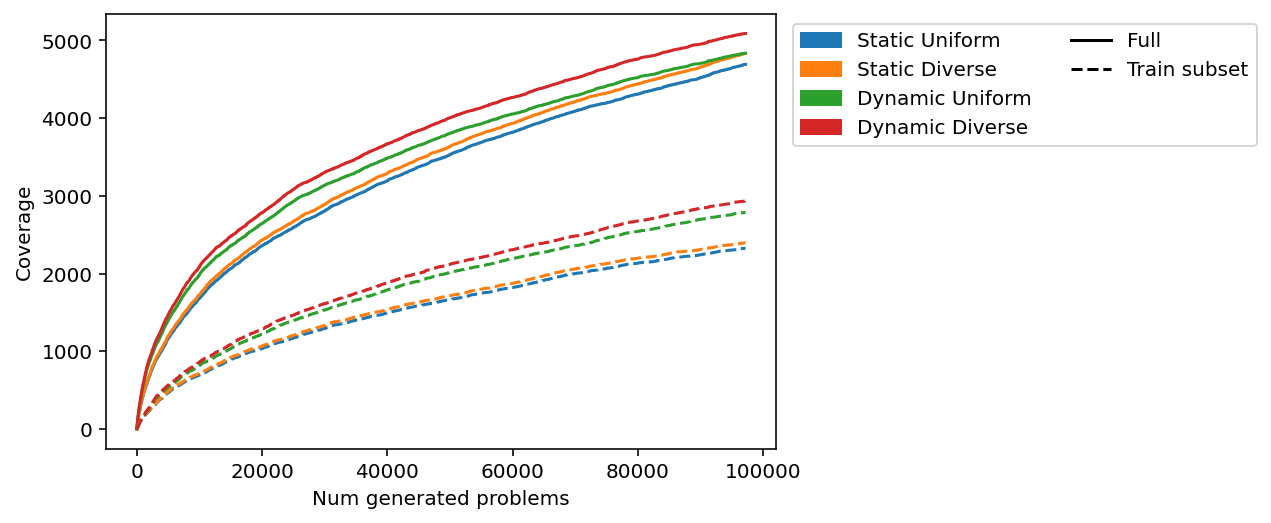}
    \caption{Coverage (number of unique skill-sets discovered) vs. number of problems generated. The QD algorithm achieves consistently higher coverage than static generation. \textbf{Note:} the train subset considers only generated problems with quality greater than $0$.
    }
    \label{fig:coverage}
\end{figure}

\ppara{Dynamic diverse (QD) methods produce the most diverse data} Figure \ref{fig:coverage} plots the diversity, as measured by number of unique problem skill-sets, of the unfiltered data archives and filtered train datasets produced by each method against the number of generated problems. All methods discover novel combinations of skills to produce a more diverse archive as the number of generated problems increases. The dynamic diverse method, combining dynamic updates to the working set with a diversity-focused partitioning of the working set, produces the most diverse data by discovering ~5000 new skill combinations after 100K problems. In contrast, the static uniform method produces the least diverse data, discovering ~4250 combinations. This gap becomes even larger when restricting to the training subset: dynamic diverse training data contains 3000 skill-sets, whereas static uniform training data contains only 2000. As a result, the training data generated by dynamic diverse methods is significantly more diverse than data generated by static uniform methods.

\ppara{Static uniform models perform best downstream} Despite the more diverse nature of dynamic diversely generated data, the static uniformly generated data leads to the best downstream models both in-distribution and OOD. This gap persists across all problem generation sizes, with the dynamic diverse method reaching a final in-distribution performance of 44\% vs. 47\% with static uniform generation. This relatively small difference in in-distribution performance is perhaps unsurprising, as our seed dataset D is already closely aligned with the test set. Generating more diverse data that covers a wider range of problems does not appear to be as beneficial for improving performance on the distribution from which $\mathcal{D}$ was generated. This underscores a critical point: selecting the \textit{right} notion of diversity when generating data depends crucially on the downstream tasks of interest.

Perhaps more surprisingly, the static uniform method also shows superior performance in the OOD setting, once again outperforming the other approaches.
The dynamic uniform approach, which iteratively updates the working set with the highest-quality mutations, comes close but still under-performs. The gap between these two methods could be due to a key difficulty the dynamic method faces, namely the reliability of the $\textsc{Quality}$ measure. We have no guarantee that the problem-solution pair $(Q, A)$ with a low solve-rate (and thus high-quality) contains logically valid reasoning. As a result, it becomes difficult to filter out high-quality but logically-invalid problems from the working set. This has the following unintended negative effect on our working set: when such a high-quality invalid sample is selected for mutation, it becomes likely to generate yet more high-quality but logically invalid data. This allows for dynamic generation methods in particular to reward hack the $\textsc{Quality}$ measure. In the next paragraph we conduct an initial investigation into this relationship, confirming that difficult but solvable problems are more likely to be invalid. However, in Section \ref{subsec:filter_analysis} we also confirm that training on higher-quality samples leads to better performance, suggesting that invalid samples can still be helpful at train time (while unhelpful during problem generation).

\begin{figure}
    \centering
    \includegraphics[width=0.48\linewidth]{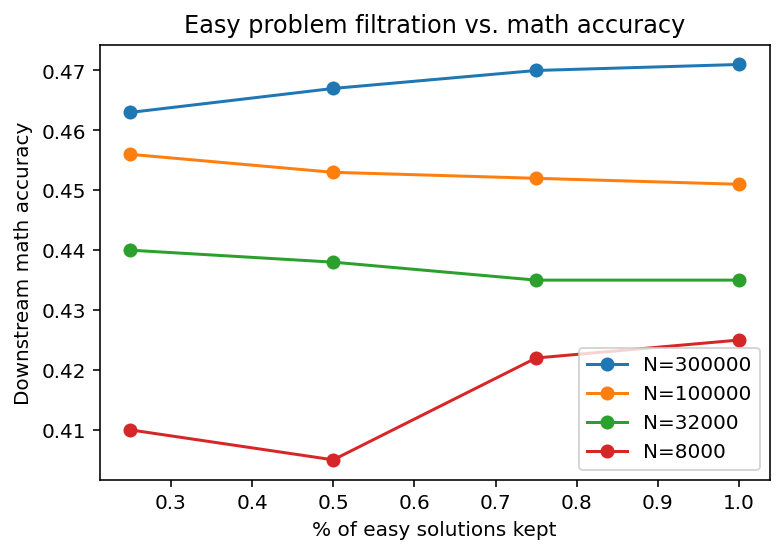}
    \caption{\% of easy verifications kept versus downstream math accuracy. Filtering out easy verifications in training positively benefits lower data regimes but negatively affects higher data regimes.}
    \label{fig:easy_filtering}
\end{figure}

\ppara{Filtering easy verifications gives mixed results} By default the construction of our training dataset in Section \ref{sec:experiments} is biased towards easier problems. This is because for a problem $Q$ with quality above the threshold $T_q$ we construct training samples $(Q, S_k)$ by pairing $Q$ with all successful verifications $S_k$. Easier problems will be verified more often (by definition) and as a result contribute more samples per problem to the training data. This results in a bias towards easy problems in the training data which may not be desirable. To account for this, we experiment with removing 75\% of the easy $(Q, S_k)$ pairs per easy $Q$. Note: we define $Q$ as \textit{easy} if $\solverate((Q, A),$ \student $) \geq 0.5$. Figure \ref{fig:easy_filtering} plots the resulting change in performance for N=100K and N=300K generated problems. Filtering the 300K generated problem set significantly reduces the size of the training dataset from 500K $(Q, S_k)$ pairs to 350K.

We find filtering has mixed results. For a smaller amount of data filtering has a positive impact, increasing performance by .075\%. However, for a larger amount of data the effect is reversed. This indicates that the performance increase from harder problems may be stable once a critical mass is reached. In contrast, with less data there are less difficult problems available and thus the effect of training on many easy problems is damaging.

\begin{figure}
    \centering
    \includegraphics[width=0.48\linewidth]{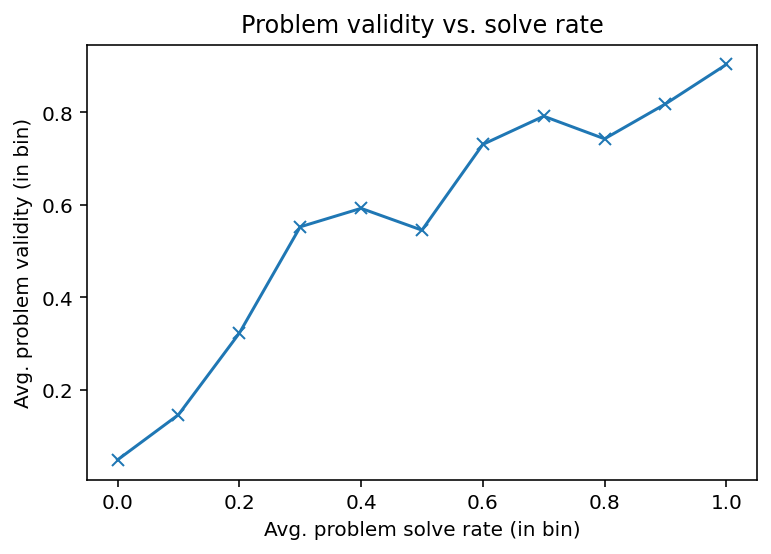}
    \caption{Average problem solve-rate versus problem validity. A higher solve-rate strongly correlates with validity, suggesting harder problems are less likely to be logically valid.}
    \label{fig:score_validity}
\end{figure}

\label{para:quality_noise}
\ppara{High-quality problems can be noisy} Our results demonstrate the solve-rate based quality measure for filtering synthetically generated problem-solution pairs positively impacts model performance. However, we are by no means guaranteed a high-quality pair $(Q, A)$ is a \textit{logically valid} pair, i.e. the problem and intended solution do not contain reasoning errors (instead we simply have $(Q, A)$ is difficult for \student to solve according to the intended solution $A$). Intuitively, we might expect problems with a higher solve-rate (and thus lower quality) to have higher chance of being valid: simply because \student consistently arrives at the intended solution. We investigate this relationship empirically by labeling the correctness of synthetic $(Q, A)$ pairs by using a SOTA reasoning model (\texttt{Gemini-2.5-flash}) to label $Q$ with alternative solutions $A'$. The final solutions of $A$ and $A'$ are then compared and used to label $(Q, A)$ as \textit{valid} if $A$, $A'$ agree and \textit{invalid} otherwise. 

In Appendix Figure \ref{fig:score_validity} we bin these labeled problems into varying solve-rate levels and plot the average validity of samples in each bin. The plot demonstrates a strong correlation between the likely validity of a problem and its solve-rate, and thus a strongly inverse relationship between validity and quality. Specifically, the harder a problem is for \student to solve, the more likely it is to be invalid. Surprisingly, as demonstrated above, this does not result in decreased fine-tuning performance when training on higher quality problems. This suggests fine-tuning to be robust to potentially high levels of invalid reasoning/noise in training data. For more investigation into the effect of the quality measure, see Section \ref{subsec:filter_analysis}.

\subsection{Results for Problem Filtering with a Fixed Training Sample Budget}
\label{subsec:filter_analysis}

\begin{figure}
    \centering
    \includegraphics[width=0.4\linewidth]{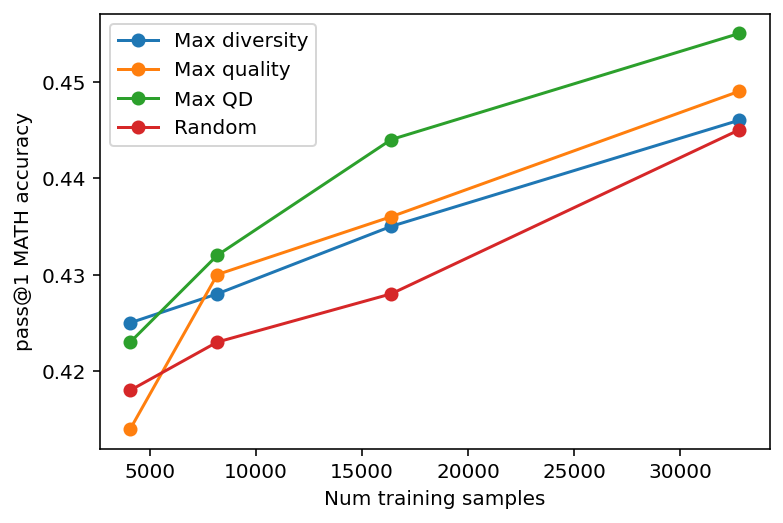}
    \includegraphics[width=0.4\linewidth]{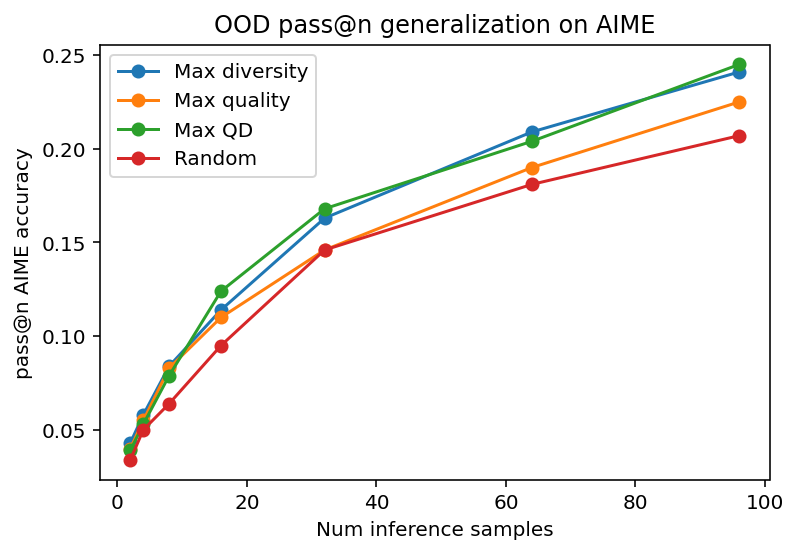}
    \caption{\textbf{Left:} In-distribution data scaling curves for various filtering strategies. On the x-axis is the number of training samples $N$ (distinct from the number of generated problems in Figure \ref{fig:scaling_results}). On the y-axis is MATH performance of fine-students. Each curve plots the performance of a different data filtering strategy. \textbf{Right:} OOD inference scaling curves for filtering strategies trained with $N=2^{15}$ samples. On the x-axis is the number of inference-time samples. On the y-axis is pass@n performance on AIME. Each curve plots the performance of a different data filtering strategy.}
    \label{fig:qd_scaling}
\end{figure}

In the previous section, we examined the downstream performance of four variants of \method{} while keeping the phase 2 filtering strategy constant.
We showed the static uniform synthetic data generation method improves over baseline SFT performance. However, the dynamic uniform method designed to jointly improve both the quality and diversity of data under-performs naive static uniform generation.

In an attempt to better understand how training data quality and diversity affect model performance, we now fix a single data generation method and explore different filtering approaches that result in different levels quality and diversity.
In particular, we conduct a series of ablations on the archive $\mathcal{A}_{\text{su}}$ of 300K problems generated via the static uniform method.

We pre-process the archive by removing all pairs with $\textsc{Quality}(Q,A) = 0$, leaving approximately 150K $(Q, A)$ pairs. We then select a single successful verifying rollout $S_Q$ for each $Q$. This gives us a train sample pool $\mathcal{D}_{\text{train}}' = \{(Q, S_Q) : (Q, \cdot) \in \archive_{\text{su}}, S_Q \text{ verifies } Q\}$ with no repeated questions. Finally, for each experiment, we will fix a sample budget $N \in \mathbb{N}$ and filter $\mathcal{D}_{\text{train}}'$ to produce a smaller training dataset with target levels of data quality and diversity. We then train \student on the subset and evaluate its in-distribution and OOD performance.

Fix a sample budget $2^{12} \leq N \leq 2^{15}$. We construct $N-$sample training mixtures in the following ways:

\begin{itemize}
    \item \textbf{Quality:} To ablate different levels of quality with the same fixed $N$, we sample training pairs $(Q, S_k)$ from $\mathcal{D}_{\text{train}}'$ from a Gaussian distribution with mean quality $m$ and standard deviation $0.1$. We sample using four different $m$ ranging from $0.2$ to $0.8$ and ensure that each selected $Q$ is unique. \textit{Max quality} refers to the subset sampled with mean quality $m = 0.8$.
    \item \textbf{Diversity:} We sample a maximally diverse $N-$sample subset by partitioning samples into their respective skill-sets/niches. We then select the subset of skill-sets $S = \{\phi^i\}$ which maximizes the number of unique skills $\phi^i_j$ in $S$. Representative samples are uniformly selected from each skill-set and added to the training subset.
    \item \textbf{QD:} We jointly optimize data quality and diversity by selecting a diverse set of skill-sets and then sampling high-quality problems within each problem skill-set.
    \item \textbf{Random:} As a baseline we uniformly randomly sample a subset of size $N$ from $\mathcal{D}_{\text{train}}'$.
\end{itemize}

\ppara{Jointly filtering for quality and diversity performs best} Figure \ref{fig:qd_scaling} shows the in-distribution and OOD performance of the filtered data subsets. We find that the mixture filtered to optimize both data quality and diversity performs best both in-distribution and OOD at nearly all training sample budgets. The best performing QD filtered model achieves 45.5\% accuracy on the MATH test set with just 32K training samples: within 1.5\% of the best static uniform model trained on 500K samples. QD filtering also consistently improves over the randomly filtered baseline, at times by up to 3\%.

\begin{figure}
    \centering
    \includegraphics[width=0.4\linewidth]{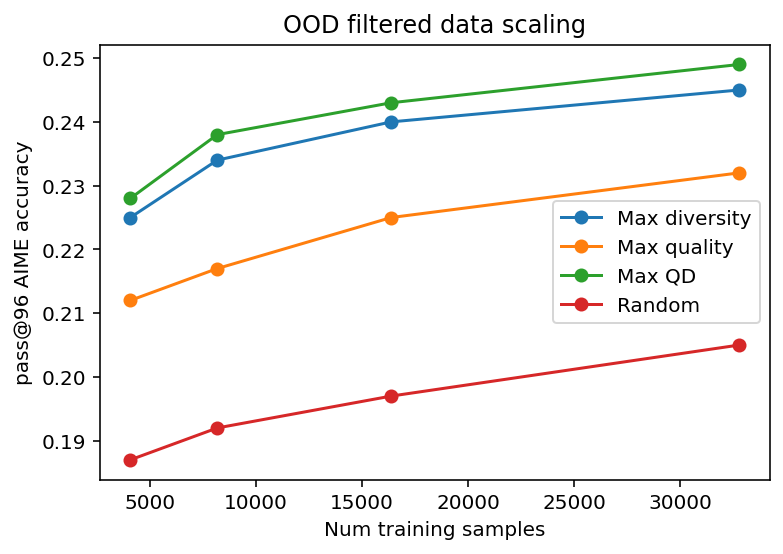}
    \includegraphics[width=0.4\linewidth]{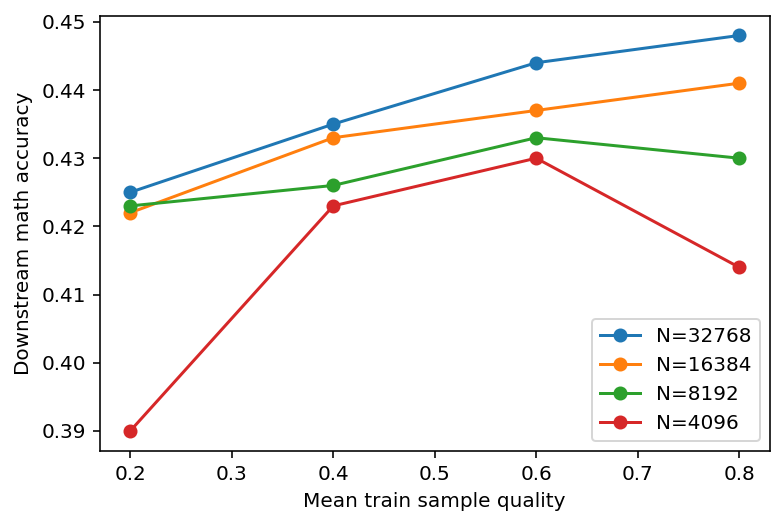}
    
    \caption{\textbf{Left:} AIME pass@96 performance of filter methods versus number of training samples. \textbf{Right:} Mean training sample quality versus MATH test performance.}
    \label{fig:aime_generalization}
\end{figure}

\ppara{Filtering for quality benefits in-distribution performance, filtering for diversity benefits OOD} Contrary to our findings in Section \ref{subsec:main_results}, here we see filtering to maximize data diversity (even without quality) leads to superior OOD performance comparable to models trained on the QD subset. With $K=96$ samples the diversely trained model improves over the random baseline by 6\% (from 19\% to 25\%). Notably, the superiority of the diversely trained model is not clear until at least $K = 16$ samples, after which the gap with Quality and Random continues to widen. This suggests that \textbf{diversely trained models may scale better with inference compute}. Finally, the left side of Figure \ref{fig:aime_generalization} illustrates the OOD performance of each method with $K=96$ inference samples against the size of training data. We find the OOD performance comparisons for each method stay consistent across different amounts of training data. However, for in-distribution performance, a model trained on only high-quality (but not diverse) data performs better. As in Section \ref{subsec:main_results}, this suggests quality is more beneficial than diversity for in-distribution performance. 

\label{para:higher_quality}
\ppara{Higher-quality data correlates with better performance} So far we have yet to confirm training on higher-quality data yields better downstream model performance. We do so by sub-sampling $\mathcal{D}_{\text{train}}'$ via a Gaussian distribution with quality mean $m$ one of $0.2,0.4,0.6,0.8$. The right side of Figure \ref{fig:aime_generalization} plots the MATH test performance of models trained on subsets with different mean levels of quality. For training sets with size greater than or equal to 8192, we see quality has a positive impact on performance. This confirms our choice of quality measure via the student \student solve-rate to be a viable proxy for data quality. When selecting $N = 2^{15}$ samples, filtering for data with a mean quality of $0.8$. vs. $0.2$ can lead to over 2.5\% absolute improvement. Note this occurs even in spite of our findings in Section \ref{subsec:main_results} showing harder problems are likely to contain logical mistakes. This suggests that \student benefits from harder problems during training even with moderate levels of noise.

\ppara{Towards more performant methods for QD-driven synthetic data generation} In Section \ref{subsec:main_results} we found that directly optimizing for higher quality samples (as done in dynamic generation methods) or more diverse samples (as done diverse methods) failed to improve over the simpler static uniform approach. Yet, our current investigation in Section \ref{subsec:filter_analysis} demonstrates benefits when directly filtering for quality and diversity given the larger dataset $\mathcal{D}_{\text{train}}'$. This suggests several avenues for improving QD-driven variants of \method{}. Firstly, the algorithms in Section \ref{subsec:main_results} are compute equalized (i.e., the same number of problems are generated) but not training sample equalized (i.e., one algorithm may generate more viable training samples than another). This translates into a difference in training sample \textit{yield-rate} where the static uniform algorithm produces viable samples for training more consistently. Secondly, the hackablity of our proposed \textit{Quality} measure becomes an important concern. Dynamic generation methods will insert high-quality but logically invalid samples into the working set, resulting in a higher percentage of high-quality but logically invalid samples in the future. In Appendix \ref{sec:perturbative_verification} we explore an attempt to detect and mitigate these samples by looking at the $\textsc{Quality}$ distributions of their children.

\subsection{Towards Recursive Self-Improvement}
\label{subsec:self-improvement}

\begin{figure}[ht]
    \centering
    \includegraphics[width=0.4\linewidth]{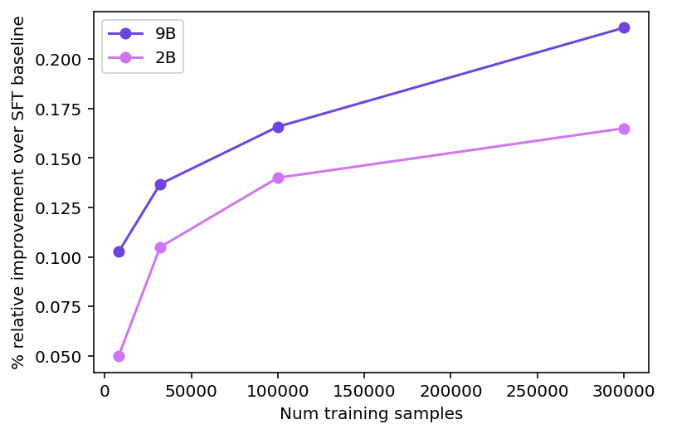}
    \includegraphics[width=0.4\linewidth]{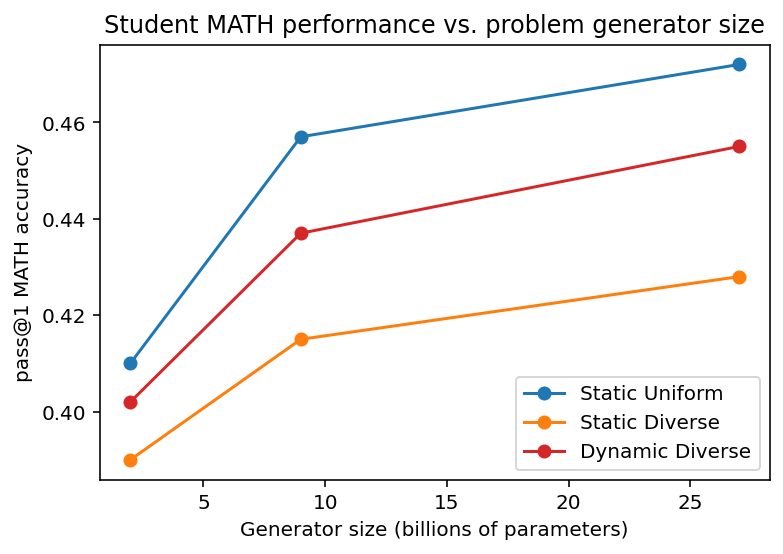}
    \caption{\textbf{Left:}\% improvement of end-to-end static self-problem generation (same model as generator and verifier). \textbf{Right:} Improvement from scaling problem generator model size. \texttt{Gemma-2-9B} acts as the student \student.}
    \label{fig:end_to_end_improvement}
\end{figure}

In our previous experiments we used $\texttt{Gemma-2-27B-it}$ as the generator \generator and a fine-tuned $\texttt{Gemma-2-9B}$ as \student. As a result, while all solutions in the training data are self-generated by the student \student, the problems are not.\footnote{Note that in most self-improvement style works the problems are fixed and thus not self-generated.} Now, we apply our methods so that the same models (\texttt{Gemma-2-2B}, \texttt{Gemma-2-9B}) act uniquely as both the generator \generator and student \student. Results for the \% relative improvement of each model over its respective SFT baseline via static uniform data generation are reported in Figure \ref{fig:end_to_end_improvement}.

\ppara{Models can self-improve by generating their own problems} After generating 300K problem-solution pairs, we find the 9B model self-improves by a relative 20\% and the 2B model self-improves by a relative 15\%. Smaller improvements persist when generating less data. On the right side of Figure \ref{fig:end_to_end_improvement} we also plot the absolute performance of the 9B student when trained using problems generated by a generator of varying size. This shows larger models generate problems more useful for the self-improvement of \student. For example, using a small 2B generator results in only 5\% improvement over the baseline. Scaling up the generator \generator adds another 2.5\% and 1.5\% improvement cumulatively. These results suggest that applying static uniform data generation to larger models benefits both from an increase in the model's problem solving ability and the model's ability to generate good questions.

\vspace{-0.4cm}
\section{Related Works}

\ppara{Synthetic data generation for reasoning} Many works have shown the benefit of synthetic data generation for reasoning by distilling from a large teacher model to a smaller student \citep{yu2024metamathbootstrapmathematicalquestions, yue2023mammothbuildingmathgeneralist,li2024mugglemathassessingimpactquery,liu2024augmentingmathwordproblems, luo2025wizardmathempoweringmathematicalreasoning}. Other works (many using RL for LLMs) generate only novel solutions to a fixed problem set \citep{havrilla2024teachinglargelanguagemodels, singh2024humandatascalingselftraining,deepseekai2025deepseekr1incentivizingreasoningcapability}. More recently, \citet{dong2025stpselfplayllmtheorem} and \citet{poesia2024learningformalmathematicsintrinsic} used small-sized LLMs to generate novel problem and solution pairs to reasoning problems in a formal environment. This differs from our work where we do not rely on any ground truth environment (e.g. Lean) to evaluate the quality of novel problems. \citet{lin2025learningsolveverifyselfplay} used small-sized LLMs to jointly generate and verify novel code problems. Related ideas to the solve-rate based quality measure used in this work have been used in open-ended reinforcement learning \citep{openendedlearningteam2021openendedlearningleadsgenerally} to prioritize efficient level sampling. Most related, \citet{pourcel2024acesgeneratingdiverseprogramming} uses a similar metric with a quality-diversity algorithm to generate difficult programming puzzles for a novel benchmark programming benchmark. In contrast, our work focuses on applying QD inspired ideas for training data generation and thoroughly ablating the effects of our quality and diversity measures on model performance.

\ppara{QD x LLMs} The number of works at the intersection of Quality-Diversity methods has been increasing rapidly over the last several years \citep{lehman2022evolutionlargemodels,bradley2023qualitydiversityaifeedback,meyerson2023language,zhang2023omni,samvelyan2024rainbow,wu2024evolutionarycomputationeralarge,chao2024match,samvelyan2024rainbow,havrilla2024surveyingeffectsqualitydiversity}. \citet{lehman2022evolutionlargemodels} were the first to utilize LLMs in evolutionary loop by evolving racing agents. \citet{bradley2023qualitydiversityaifeedback} utilized AI feedback to iteratively synthesize high-quality poetry. \citet{zhang2023omni} utilized powerful LLMs to generate a diverse set of RL environments for training open-ended agents. See \citet{havrilla2024surveyingeffectsqualitydiversity} for an in-depth review of the intersection of QD with LLMs for synthetic data generation.

\vspace{-0.25cm}
\section{Conclusion and Limitations}
\label{sec:conclusion}

\vspace{-0.25cm}
In this work, we presented \method{}, a new approach for generating high-quality and diverse synthetic math problem by optimizing both data quality and diversity. 
We find that training on the resulting data with \method{} gave absolute improvements of up to 9\% over an SFT baseline and scaled with both the size of the problem generator and the amount of generated problem data. Further, we conducted thorough ablations into the effects of data quality and diversity, finding that training on high-quality data leads to better in-distribution generalization and training on more diverse data can lead to better OOD generalization. Future work might address some of the current method's limitations by focusing on designing better QD inspired data generation algorithms with improved training sample yield rates and mitigating over-optimization of our quality measure.

\bibliographystyle{plainnat}
\bibliography{refs}


\appendix

\section{Training Hyperparameters}
\label{sec:hparams}

We fine-tune the pre-trained version of \student for a single epoch on the resulting $\mathcal{D}_{\text{train}}$. We use learning rate $lr = 2e-6$ with a cosine schedule decaying to $2e-7$ with a batch size of 16. Training is done on a slice of 8x8 TPUv4s.

\section{Measuring Noisy Problem Quality via \textbf{Perturbative Verification}}
\label{sec:perturbative_verification}

\begin{figure}
    \centering
\includegraphics[width=0.32\linewidth]{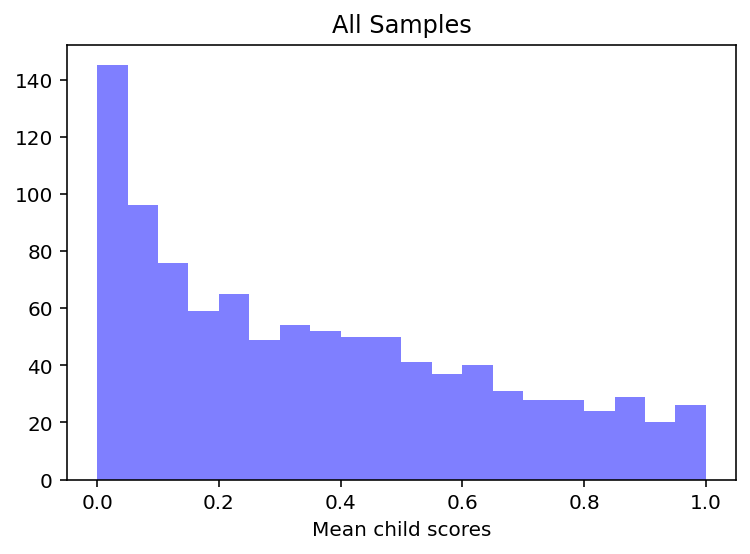}
\includegraphics[width=0.32\linewidth]{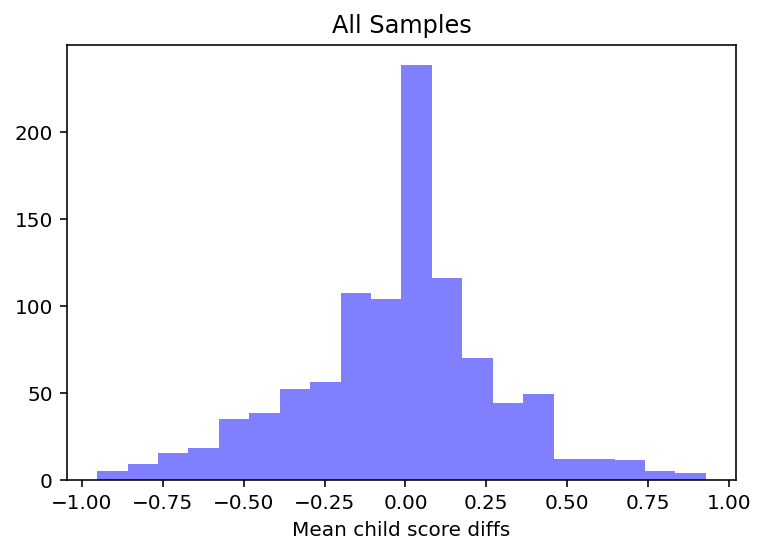}
\includegraphics[width=0.32\linewidth]{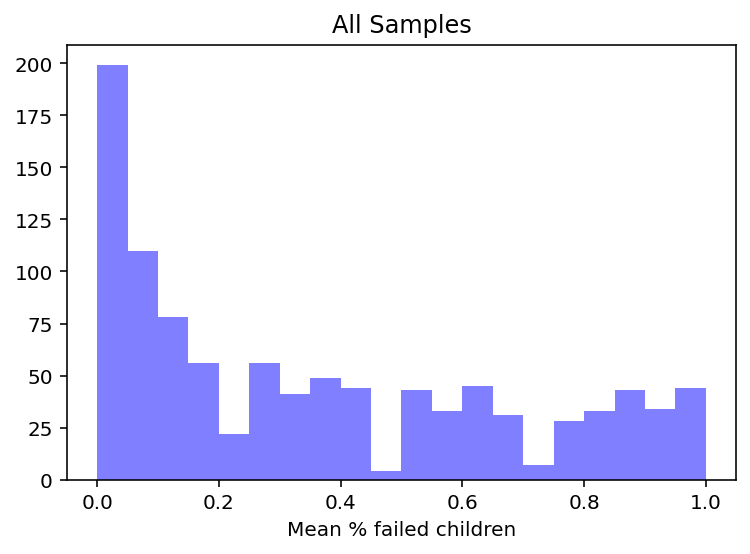}

\includegraphics[width=0.32\linewidth]{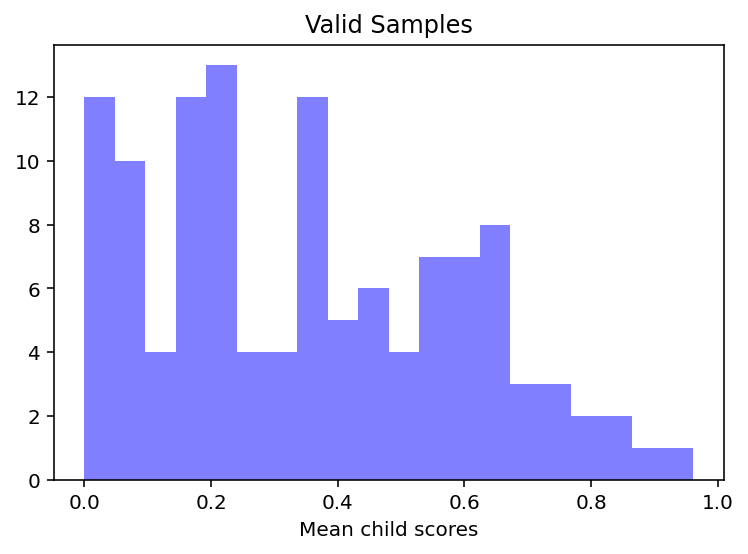}
\includegraphics[width=0.32\linewidth]{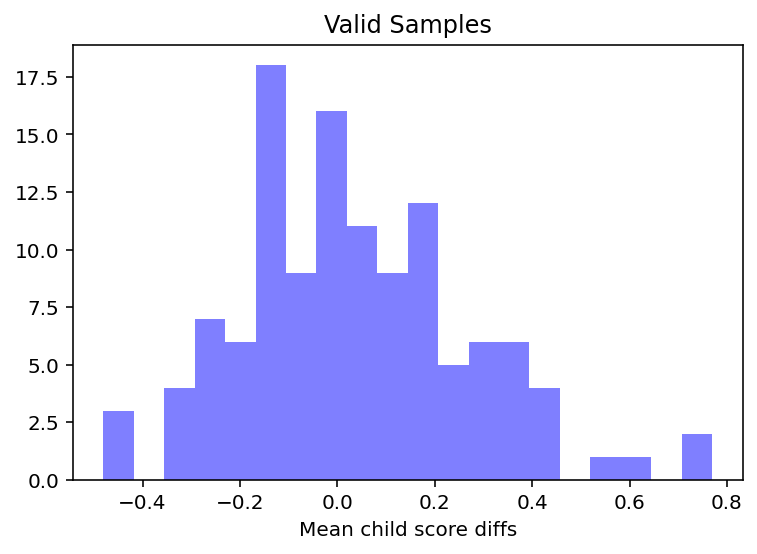}
\includegraphics[width=0.32\linewidth]{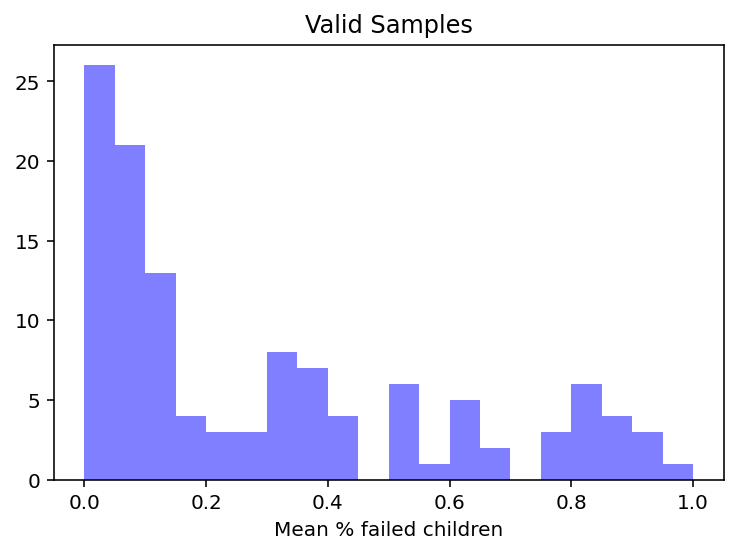}

\includegraphics[width=0.32\linewidth]{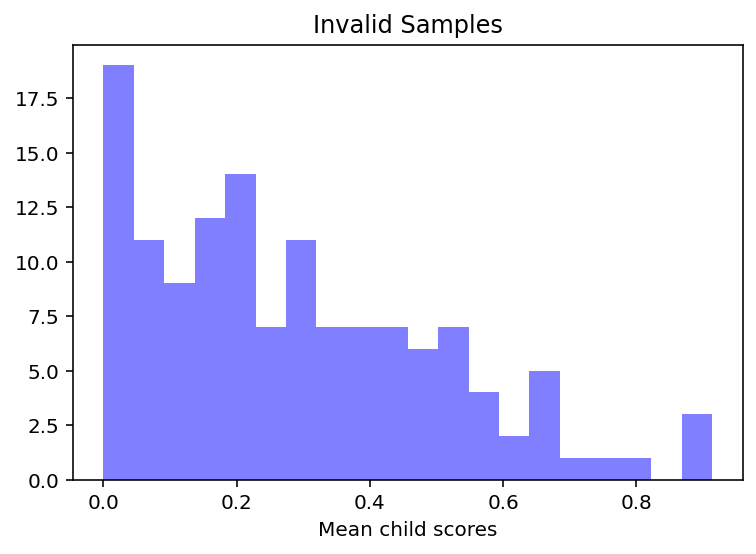}
\includegraphics[width=0.32\linewidth]{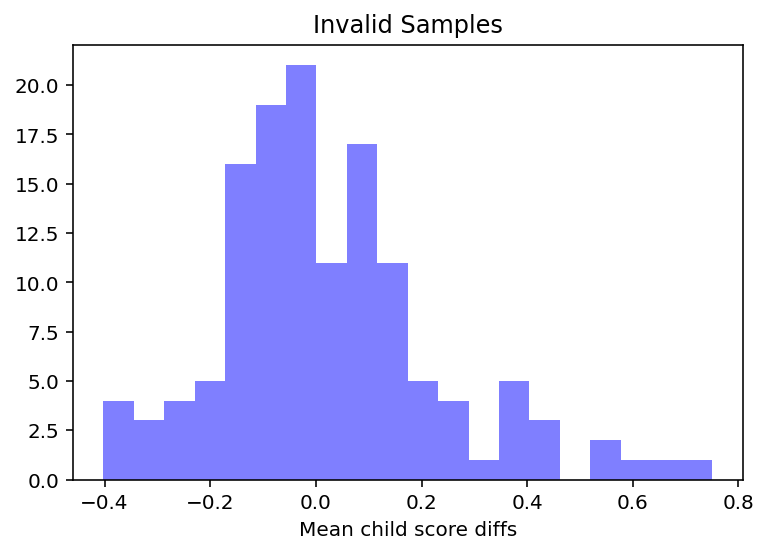}
\includegraphics[width=0.32\linewidth]{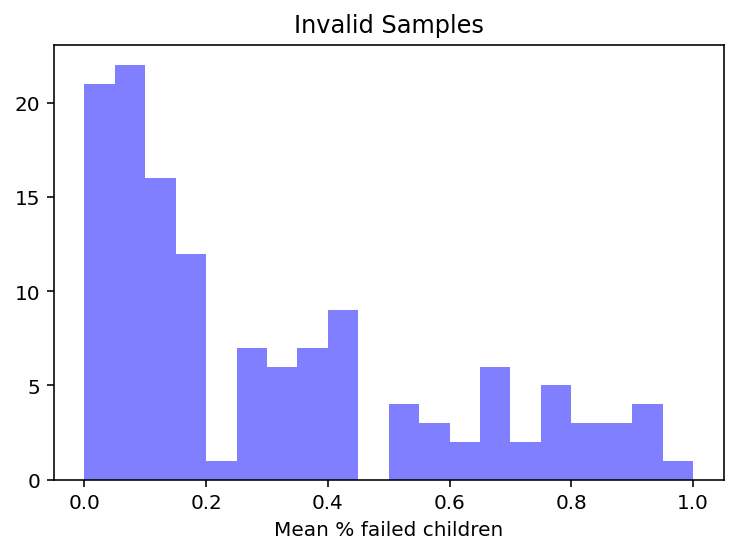}

    \caption{Child score distributions for valid and invalid samples.}
    \label{fig:valid_invalid_score_dists}
\end{figure}

In Section \ref{sec:methods} we use a reference model \student to introduce the solve-rate as a proxy to measure a problem-solution pair's quality/difficulty. Figure \ref{fig:easy_filtering} shows that harder problems (those successfully verified less often than easy problems) more often contain logical inaccuracies leading to an incorrect solution. As a result, many $(Q, A)$ pairs evaluated as high-quality contain mistakes depending in some way on the behavior of the verifier \student. This makes the score rate a \textit{noisy} proxy of quality and leads to the following question: how can we efficiently filter out $(Q, A)$ pairs with low solve-rate which are logically invalid? One potential solution is to evaluate the validity of a sample $(Q, A)$ via the measured quality of its downstream mutations/children $(Q_1', A_1'),...,(Q_n', A_n')$.

The intuition goes as follows: suppose we have two pairs $(Q_V, A_V)$, $(Q_I, A_I)$ , with identical corresponding quality scores $q = q_V = q_I$, which are valid/invalid respectively. Now suppose we sample $n$ mutations of both pairs using generator \generator to produce child sets $(Q_V^{1}, A_V^{1}),...,(Q_V^{n},...,A_V^{n})$ and $(Q_I^{1}, A_I^{1}),...,(Q_I^{n},...,A_I^{n})$ with corresponding quality scores $q_V^1,...,q_V^n$ and $q_I^1,...,q_I^n$. In order to identify the invalid pair $(Q_I, A_I)$ we might hope that \textbf{the invalid sample generates more invalid children than the valid sample}. We investigate two possible measurements to detect whether this is the case: 1) the mean difference of the scores of a parent with its children 2) the \% of children of a parent with quality $q = 0$. Formally, we define the mean difference as $\frac{\sum_{i=1}^n(q - q^i)}{n}$ and the child failure rate as $\frac{\left| \{q^i : q^i = 0\} \right|}{n}$. To examine the quality distributions of children of valid and invalid samples we take the 1000 annotated $(Q, A)$ pairs in Section \ref{sec:experiments} and generate $n = 16$ mutations of each. We then compute quality scores for each mutation via the score rate based on $\texttt{Gemma-2-9b}$ as \student. 

Figure \ref{fig:valid_invalid_score_dists} plots the distribution of mean child scores, differences between mean child score and parent score, and \% of children with a score of $0$. The distribution of mean child scores for all samples is left skewed towards $0$ since the majority of parents have a score of $0$. The mean scores of children concentrate around their parent's score with gaussian-like decay in the tails. When plotting the distributions for valid/invalid samples we keep only the samples with solve-rate between $l = 0.1$ and $u=0.5$. This is done i) to remove samples with a score of $0$ and ii) remove easy samples skewing the scores of valid problems (since easy problems are more likely to be valid). Surprisingly, after filtering in this way, we find valid and invalid samples have similar distributions of child scores. Both have a mean child parent score difference of $0.03$ and around 28\% children with $0$ score. This suggests a different mutation mechanism more sensitive to the validity/invalidity of parents may be needed to find a difference between respective child scores or a more reliable oracle may be needed.

\section{The Impact of Seed-Dataset Size}
\label{sec:seed_size}

To investigate whether the seed dataset size impacts the relative performances of static uniform and dynamic diverse data generation algorithms we randomly sample small seed dataset with 700 samples from $\mathcal{D}$ to produce $\mathcal{D}_{\text{small}}$. We run both algorithms to produce 10,000 samples each and fine-tune \student on the resulting data. We find the model trained on static uniformly generated data gets 40\% on MATH test whereas the model trained on dynamic diverse data gets 39\% on MATH test.

\section{A Skill-Unbounded QD Algorithm}
\label{sec:skill_unbounded}

In Section \ref{sec:experiments} when implementing the dynamic diverse algorithm we limit the diversity of problems by only considering the top $M = 100$ most commonly occurring skills. Additionally, when selecting samples for mutation, we uniformly sample skill-sets. This selection procedures does not take into account inter-skill-set similarities where one skill (e.g. $\texttt{Algebra}$) may be repeated across many skill-sets. To promote even more sample diversity we propose the following modifications to the dynamic diverse data generation algorithm:

\begin{itemize}
    \item We allow for an unbounded set of possible skills generated by the skill-classifier LLM. However, we still restrict the skill-set description of problem to the three most relevant skills.
    \item We sample a set of skill-sets for mutation by solving the skill-sets optimization problem $\text{argmax}_S \textsc{DIV}(S)$ where the diversity measure $\textsc{DIV}$ of a collection of skill-sets $S = \{\phi_1,...,\phi_n\}$ is the number of unique skills $\phi^i$ in $S$. 
\end{itemize}

We use this algorithm to generate 100,000 synthetic problem-solution pairs and train a model on the resulting data. The model gets 43\% accuracy on MATH test and 21\% accuracy on AIME with $K = 96$ samples. This is slightly worse than the unmodified dynamic diverse method.

\section{Solve-rate Distributions of Synthetically Generated Problems}

\begin{figure}
    \centering
    \includegraphics[width=0.48\linewidth]{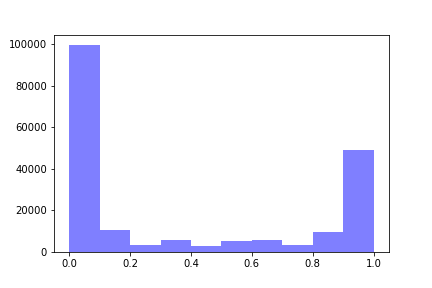}
    \includegraphics[width=0.48\linewidth]{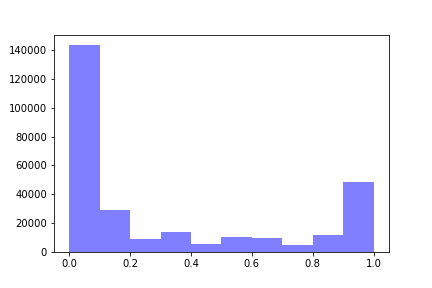}
    \caption{\textbf{Left:} Solve-rate distribution of statically generated problems. \textbf{Right:} Solve-rate distribution of QD generated problems. The vast majority of generated problems are either too hard (\textbf{SR} = 0) or too easy (\textbf{SR} = 1).}
    \label{fig:score_dists}
\end{figure}

See Figure \ref{fig:score_dists} for a histogram showing the solve-rate distributions of static uniformly and dynamic diversely generated problems. The vast majority are never solved by \student $(\textsc{SolveRate} = 0)$ or are too easy $(\textsc{SolveRate} = 1)$.

\section{Prompts}
\label{sec:prompts}

\begin{tcolorbox}[colback=yellow!30, colframe=blue!40!black, title=Mutation Prompt]
You are tasked with generating a mutation conditioned on a set of input problems. You will be shown the problems below.

\{problem\}

\{solution\}

Now generate a novel problem and solution. \
Enclose the problem in <problem>...</problem> tags and the solution in <solution>...</solution> tags.
Make sure to include the intended final answer in the solution enclosed in the \boxed{...} latex style. \
If there are multiple numerical answers, write them as a comma separated list \boxed{(n1, n2,...)}.
\end{tcolorbox}

\begin{tcolorbox}[colback=yellow!30, colframe=blue!40!black, title=Skill Classification Prompt]
You will be a shown a reasoning problem below and solution below. Your job is to list the relevant skills/lemmas used in solving the problem. \
Enclose all skills in a comma separated list enclosed in the tags <skills>...</skills>. For example, if the problem is solved using a combination of \
arithmetic and pigeonhole-principle, write <skills>arithmetic,pigeonhole-principle</skills>. \
Only include up to \{k\} relevant skills.

\{problem\}

\{solution\}
\end{tcolorbox}

\section{Synthetic problem samples}
\label{sec:synthetic_problem_samples}

\begin{table}[ht]
\centering
\caption{Hard, valid sample ($\textbf{SR} = 0.25$).}
\vspace{2mm}
\begin{tabularx}{\textwidth}{p{0.11\textwidth}X}
\toprule
\parboxc{Problem} & \parboxc{
What is the smallest base $b$ for which the base $b$ representation of $1000_{10}$ requires exactly 4 digits?
} \\
\midrule
\parboxc{Solution} & \parboxc{
For a base $b$ representation of $1000_{10}$ to have exactly 4 digits, the largest power of $b$ that is less than $1000$ must be $3$. Therefore, we have the requirement that $b^3 \\le 1000 < b^4$.  We then realize that $b=6$ satisfies this requirement since $6^3 < 1000 < 6^4$. We also realize that this is the smallest possible value of $b$ since if $b$ were less than $6$, $b^4$ would be less than $1000$. Thus, our only solution is $b = \boxed{6}$.
} \\
\bottomrule
\end{tabularx}
\end{table}

\begin{table}[ht]
\centering
\caption{Easy, invalid sample ($\textbf{SR} = 0.5625$).}
\vspace{2mm}
\begin{tabularx}{\textwidth}{p{0.11\textwidth}X}
\toprule
\parboxc{Problem} & \parboxc{
We have that $2x^2f'(x) - 3xg(x) = h'(x)$ where $f(x),$ $g(x),$ and $h(x)$ are all polynomials in $x$ and $f'(x)$ and $h'(x)$ denote the derivatives of $f(x)$ and $h(x)$ respectively. If the degree of $h(x)$ is $6,$ what is the minimum possible degree of $g(x)$?
} \\
\midrule
\parboxc{Solution} & \parboxc{
Since $h'(x)$ has degree $5$ (one less than the degree of $h(x)$), the degree of the left-hand side of the equation must also be $5$. The term involving $f'(x)$ will have a degree that is one less than the degree of $f(x)$.  Therefore, to achieve a degree of 5 on the left-hand side, the term $-3xg(x)$ must have a degree of 5. This means the minimum degree of $g(x)$ is $\boxed{4}$.
} \\
\bottomrule
\end{tabularx}
\end{table}

\begin{table}[ht]
\centering
\caption{Hard, invalid sample ($\textbf{SR} = 0.125$).}
\vspace{2mm}
\begin{tabularx}{\textwidth}{p{0.11\textwidth}X}
\toprule
\parboxc{Problem} & \parboxc{
Find the quadratic polynomial, with real coefficients, which has $-2i$ as a root, and where the constant term is 5.
} \\
\midrule
\parboxc{Solution} & 
\parboxc{
Since the polynomial has real coefficients, the other root must be $2i.$  Thus, the polynomial is
\begin{align*}(x - (-2i))(x - 2i) &= (x + 2i)(x - 2i)\\
&= x^2 - (2i)^2 \\
&= x^2 + 4 \\
&= \boxed{x^2 + 4}.\\
\text{We then add 5 to get a constant term of 5:} & \\
x^2 + 4 + 5 &= \boxed{x^2 + 9}.
\end{align*}
} \\
\bottomrule
\end{tabularx}
\end{table}

\end{document}